\documentclass[conference]{IEEEtran}
\IEEEoverridecommandlockouts
\usepackage{cite}
\usepackage{amsmath,amssymb,amsfonts}
\usepackage{algorithmic}
\usepackage{graphicx}
\usepackage{textcomp}
\usepackage{algorithm}
\usepackage{amsfonts}
\usepackage{cite}
\usepackage{graphicx}
\usepackage{amssymb}
\usepackage{caption}
\usepackage{epsfig}
\usepackage{subfigure}
\usepackage{amsmath}
\usepackage{array}
\usepackage{color}
\usepackage{amsmath}
\usepackage{longtable}
\usepackage{algorithmic}
\usepackage{algorithm}
\usepackage{cases}
\usepackage{mathrsfs}
\usepackage{psfrag}
\usepackage{url}
\usepackage{setspace}
\usepackage{float}

\usepackage[dvipsnames, svgnames, x11names]{xcolor} 
\usepackage{ulem}
\usepackage{footnote}
\usepackage{etoolbox}
\makeatletter
\patchcmd{\@makecaption}
  {\scshape}
  {}
  {}
  {}
\makeatletter
\patchcmd{\@makecaption}
  {\\}
  {.\ }
  {}
  {}
\makeatother

\IEEEoverridecommandlockouts

\def\BibTeX{{\rm B\kern-.05em{\sc i\kern-.025em b}\kern-.08em
    T\kern-.1667em\lower.7ex\hbox{E}\kern-.125emX}}
\newtheorem{Thm}{Theorem}

\newtheorem{Asump}{Assumption}
\newtheorem{Prob}{Problem}
\newtheorem{Proof}{Proof}

\newtheorem{Rmk}{Remark}

\newcommand{\setNbar}{\bar{\mathcal N}}

\begin{document}

\title{Optimization-Based GenQSGD for Federated Edge Learning}
\author{\IEEEauthorblockN{Yangchen Li}\IEEEauthorblockA{Shanghai Jiao Tong University, China\\
liyangchen@sjtu.edu.cn}\thanks{This work was supported by the Natural Science Foundation of Shanghai under Grant 20ZR1425300.}
\and
\IEEEauthorblockN{Ying Cui}
\IEEEauthorblockA{Shanghai Jiao Tong University, China\\
cuiying@sjtu.edu.cn}
\and
\IEEEauthorblockN{Vincent Lau}
\IEEEauthorblockA{HKUST, Hong Kong\\
eeknlau@ee.ust.hk}
}
\maketitle

\begin{abstract}
Optimal algorithm design for federated learning (FL) remains an open problem. This paper explores the full potential of FL in practical edge computing systems where workers may have different computation and communication capabilities, and quantized intermediate model updates are sent between the server and workers. First, we present a general quantized parallel mini-batch stochastic gradient descent (SGD) algorithm for FL, namely GenQSGD, which is parameterized by the number of global iterations, the numbers of local iterations at all workers, and the mini-batch size. We also analyze its convergence error for any choice of the algorithm parameters. Then, we optimize the algorithm parameters to minimize the energy cost under the time constraint and convergence error constraint. The optimization problem is a challenging non-convex problem with non-differentiable constraint functions. We propose an iterative algorithm to obtain a KKT point using advanced optimization techniques. Numerical results demonstrate the significant gains of GenQSGD over existing FL algorithms and reveal the importance of optimally designing FL algorithms.
\end{abstract}


\setcounter{page}{1}

\section{Introduction}
With the development of mobile Internet and Internet of Things (IoT), massive amounts of data are generated at the edge of wireless systems and stored in a distributed manner.
Meanwhile, many applications involve machine learning with distributed databases.
It may be impossible or undesirable to upload distributed databases to a central server due to energy and bandwidth limitations or privacy concerns.
Recent years have witnessed the growing interest in federated learning (FL), which can successfully protect data privacy for privacy-sensitive applications and improve communication efficiency.

Typical FL algorithms include Federated Averaging (FedAvg) \cite{FedAvg}, Parallel Restarted SGD (PR-SGD) \cite{YuHao}, and Federated learning algorithm with Periodic Averaging and Quantization (FedPAQ) \cite{FedPAQ}.
In these algorithms, the server periodically updates the global model by aggregating and averaging the local models trained and uploaded by the workers.
In FedAvg, within a global iteration, each participated worker utilizes all its local samples, and all workers conduct the same number of local iterations.
In PR-SGD, each worker utilizes only one local sample in each local iteration.
In \cite{YuHao}, the authors analyze the convergence of PR-SGD under the assumption that the server and workers send accurate model updates, which may contain a large number of information bits.
It is worth noting that the convergence of a mini-batch SGD-based FL algorithm with an arbitrary choice of algorithm parameters is still unknown.
In \cite{FedPAQ}, the authors characterize the convergence of FedPAQ assuming that only workers can send quantized model updates with a limited number of information bits, and all workers execute the same number of local iterations.
The algorithm parameters of FedAvg, PR-SGD, and FedPAQ are chosen empirically and experimentally.
Optimizing the algorithm parameters may significantly improve the convergence and reduce communication costs.
Realizing this key advantage, the authors in \cite{AlgDes1-1} consider the optimization of the algorithm parameters to minimize the convergence error of a mini-batch SGD algorithm for a convex machine learning problem. Unfortunately, only a heuristic solution is obtained.
A rigorous optimization framework for optimizing the algorithm parameters of a general federated learning algorithm is highly desirable.
Besides, some existing works \cite{ResAlls3-1,ResAlls3-2,ResAlls4-1} focus on optimal resource allocation when implementing FL algorithms with fixed parameters.
Such optimal resource allocation is out of the scope of this paper.

This paper considers a practical edge computing system where workers may have different computation and communication capabilities, and quantized intermediate model updates are sent between the server and workers.
To explore the full potential of FL in such an edge computing system, we first present a general quantized parallel mini-batch SGD algorithm, namely GenQSGD, which is parameterized by the number of global iterations, the numbers of local iterations at all workers, and the mini-batch size.
We also characterize three core performance metrics for GenQSGD, i.e., the time cost, energy cost, and convergence error, in terms of the algorithm parameters, system parameters, and parameters of machine learning problem.
To achieve an optimal tradeoff among the three metrics, we formulate the minimization of the energy cost with respect to the algorithm parameters under the time constraint and convergence error constraint.
The optimization problem is a challenging non-convex problem with non-differentiable constraint functions.
We propose an iterative algorithm using general inner approximation (GIA) \cite{GIA} and the tricks for solving complementary geometric programming (CGP) \cite{CGP}.
We also characterize its convergence to a KKT point.
Numerical results demonstrate the significant gains of GenQSGD over existing FL algorithms and reveal the importance of optimally designing FL algorithms.
To the best of our knowledge, this is the first work that provides a rigorous optimization framework for designing an FL algorithm according to the parameters of the underlying edge computing system and machine learning problem.
%

{\bf{Notation}}:
We represent vectors by boldface letters (e.g., $\mathbf x$),
scalar constants by non-boldface letters (e.g., $x$ or $X$),
and sets by calligraphic letters (e.g., $\mathcal X$).
$\|\cdot\|_2$ represents the $l_2$-norm.
$\mathbb E[\cdot]$ represents the expectation, and $\mathbb I[\cdot]$ represents the indicator function.
The set of real numbers is denoted by $\mathbb R$,
and the set of positive integers is denoted by $\mathbb Z_+$.
\section{System Model}
We consider an edge computing system, consisting of one server and $N$ workers, which are connected via wireless links.
Let $0$ and $\mathcal N\triangleq\{1,2,\cdots,N\}$ denote the server index and the set of worker indices, respectively.
For ease of exposition, we also denote $\setNbar\triangleq\{0\}\cup\mathcal N$.
We assume that each worker $n\in\mathcal N$ holds $I_n$ samples,
denoted by $\xi_i,i\in\mathcal I\triangleq\{1,\cdots,I\}$, which can be viewed as realizations of a random variable, denoted by $\zeta_n$.
The server and $N$ workers aim to collaboratively train a global model based on the local data stored on the $N$ workers, by solving a machine learning problem.
The global model is parameterized by a $D$-dimensional vector ${\bf x}\in \mathbb R^{D}$.
Specifically, for a given $\mathbf x\in\mathbb R^D$, define the loss incurred by $\zeta_n$ as $F({\mathbf x};\zeta_n)$ and define the expected loss as $f_n({\mathbf x})\triangleq\mathbb E\left[F({\mathbf x};\zeta_n)\right]$, with the expectation taken with respect to the distribution of $\zeta_n$, for all $n\in\mathcal N$.
Then, the expected risk function $f:\mathbb R^{D}\rightarrow\mathbb R$ of the model parameters ${\mathbf x}\in\mathbb R^{D}$ is defined as:
\begin{align}\label{eq:expected_loss}
f({\mathbf x})\triangleq\frac{1}{N}\sum_{n\in\mathcal N}f_n({\mathbf x}).
\end{align}
To be general, we do not assume $f({\bf x})$ to be convex.
Our goal is to minimize the expected risk function with respect to the model parameters ${\bf x}$ in the edge computing system.
\begin{Prob}[Machine Learning Problem]\label{Prob1}
\begin{align}\label{Obj_ML}
f^*\triangleq\min_{{\mathbf x}}f({\bf x}),
\end{align}
where $f({\bf x})$ is given by \eqref{eq:expected_loss}.
\end{Prob}

The server and $N$ workers all have computing and communication capabilities.
Let $F_0$ and $F_n$ denote the CPU frequencies (cycles/s) of the server and worker $n\in\mathcal N$, respectively.
Let $p_0$ and $p_n$ denote the transmission powers of the server and worker $n\in\mathcal N$, respectively.
In the process of collaborative training, the server multicasts a message to the $N$ workers over the whole frequency band at an average rate $r_0$ (b/s),
and the $N$ workers transmit their messages to the server using Frequency Division Multiple Access (FDMA).
The average transmission rate of worker $n\in\mathcal N$ is $r_n$ (b/s).

The server and all $N$ workers employ quantization before transmitting messages.
We consider a general random quantizer, parameterized by a tunable parameter $s\in\mathbb Z_+$, which corresponds to the number of quantization levels.\footnote{Usually, $s$ is the number of quantization levels or its increasing function.}
The quantization function, denoted by $\mathbf Q(\cdot;s):\mathbb R^D\rightarrow\mathbb R^D$, satisfies the following assumptions \cite{FedPAQ}.
\begin{Asump}[Assumptions on Random Quantization]\label{Asump:Quantization}
For all $\mathbf y\in\mathbb R^D$ and $s\in\mathbb Z_+$, $\mathbf Q(\cdot;s)$ satisfies:
(i) $\mathbb E\left[{\mathbf Q}(\mathbf y;s)\right]=\mathbf y$ (unbiasedness),
(ii) $\mathbb E\left[\left\|\mathbf Q(\mathbf y;s)-\mathbf y\right\|_2^2\right]\leq q_s\left\|\mathbf y\right\|_2^2$ (bounded variance),
for some constant $q_s>0$.
\end{Asump}

The number of bits to specify the quantized vector ${\mathbf Q}(\mathbf y;s)$ of an input vector $\mathbf y$, i.e., represent the input vector $\mathbf y$, is denoted by $M_s$ (bits).\footnote{Note that $q_s$ and $M_s$ depend on $s$.}
We use $s_0\in\mathbb Z_+$ and $s_n\in\mathbb Z_+$ to denote the random quantization parameters for the server and worker $n\in\mathcal N$, respectively.
Denote $\mathbf s\triangleq\left(s_n\right)_{n\in\setNbar}\in\mathbb Z_+^{N+1}$.

\begin{Rmk}[General Edge Computing System]
The considered edge computing system is general in the sense that the system parameters, e.g., $F_n,p_n,r_n,s_n \left(q_{s_n}, M_{s_n}\right),n\in\setNbar$, can be different.
\end{Rmk}

\section{Algorithm Description and Convergence Analysis for GenQSGD}\label{Sec:Alg}
In this section, we first present a general quantized parallel mini-batch SGD algorithm for solving Problem~\ref{Prob1} in the edge computing system.
Then, we analyze its convergence.

\subsection{Algorithm Description}
The proposed  general quantized parallel mini-batch SGD algorithm, namely, GenQSGD, is parameterized by $\mathbf K\triangleq\left(K_n\right)_{n\in\setNbar}\in\mathbb Z_+^{N+1}$ and $B\in\mathbb Z_+$.
Specifically, $K_0$ represents the number of global iterations,
$K_n$ represents the number of local iterations executed by worker $n\in\mathcal N$ within one global iteration,
and $B$ represents the local mini-batch size used for local iterations at each worker.
Denote $\mathcal K_0\triangleq\{1,2,\cdots,K_0\}$ and $\mathcal K_n\triangleq\{1,2,\cdots,K_n\}$ as the sets of global iteration indices and local iteration indices for worker $n\in\mathcal N$, respectively.
For all $k_0\in\mathcal K_0$, let $\hat{\mathbf x}^{(k_0)}\in\mathbb R^D$ denote the global model recovered by all $N$ workers at the beginning of the $k_0$-th global iteration.
For all $k_0\in\mathcal K_0$, let $\Delta\hat{\mathbf x}^{(k_0)}\in\mathbb R^D$ denote the average of the quantized overall local model updates at the $k_0$-th global iteration.
For all $k_0\in\mathcal K_0$ and $k_n\in\mathcal K_n$, let $\mathbf x_n^{(k_0,k_n)}\in\mathbb R^D$ denote the local model of worker $n\in\mathcal N$ at the $k_n$-th local iteration within the $k_0$-th global iteration.
Let $\gamma$ denote the constant step size used in the local iterations.\footnote{For ease of exposition, we consider constant step size in this paper as in \cite{FedPAQ,YuHao}.
The convergence analysis and optimization results in this paper can be readily extended to other step size rules, such as the exponential and diminishing step size rules.}
The details of the proposed GenQSGD are summarized in Algorithm~\ref{Alg:GenQSGD}.

\begin{Rmk}[Generality of GenQSGD]
GenQSGD is general in the sense that $\mathbf K$ and $B$ can be flexibly chosen,
and GenQSGD can be applied for any given quantization parameters $\mathbf s$.
GenQSGD includes some existing algorithms as special cases.
For the sake of discussion,
we let $s_n=\infty, n\in\setNbar$ present the case without quantization.
In particular, GenQSGD with $K_n=m\frac{I_n}{B},m\in\mathbb Z_+,n\in\mathcal N$ for $s_n=\infty,n\in\setNbar$ reduces to FedAvg \cite{FedAvg}, which does not consider quantization at the server and workers and requires each worker to make a number of training passes over its local samples within each global iteration.
GenQSGD with $B=1$ for $s_n=\infty,n\in\setNbar$ is identical to PR-SGD \cite{YuHao}, which does not consider quantization at the server and workers and utilises only one local sample in each local iteration.
GenQSGD with $K_1=K_2=\cdots=K_N$ for $s_0=\infty$ is the same as FedPAQ \cite{FedPAQ} (supposing that all workers are selected in each global iteration), which does not consider quantization at the server and requires the same number of local iterations at all workers.
\end{Rmk}

\begin{algorithm}[t]
\caption{GenQSGD}
\label{Alg:GenQSGD}
{\bf{Input:}} $\mathbf K\in\mathbb Z_+^{N+1}$ and $B\in\mathbb Z_+$.\\
{\bf{Output:}} $\mathbf x^*\left(\mathbf K,B\right)$.
\begin{algorithmic}[1]
\STATE {\bf{Initialize:}} The server generates $\mathbf x_0^{(0)}$, sets $\Delta\hat{\mathbf x}^{(0)}=\mathbf x_0^{(0)}$, and sends $\mathbf Q(\Delta\hat{\mathbf x}^{(0)};s_0)$ to all $N$ workers.
The $N$ workers set $\hat{\mathbf x}^{(0)}=0$.
\FOR {$k_0=1,2,\cdots,K_0$}
    \FOR {worker $n\in{\mathcal N}$}
        \STATE Compute $\hat{\mathbf x}^{(k_0)}:=\hat{\mathbf x}^{(k_0-1)}+\mathbf Q(\Delta\hat{\mathbf x}^{(k_0-1)};s_0)$,
        and set ${\mathbf x}_n^{(k_0,0)}=\hat{\mathbf x}^{(k_0)}$.
        \FOR {$k_n=1,2,\cdots,K_n$}
            \STATE Randomly select a mini-batch $\mathcal B_n^{(k_0,k_n)}\subseteq\mathcal I_n$ and update ${\bf x}_n^{(k_0,k_n)}$ according to: ${\mathbf x}_n^{(k_0,k_n)}$$:={\bf x}_n^{(k_0,k_n-1)}$$-\gamma\frac{1}{B}\sum_{\xi_i\in\mathcal B_n^{(k_0,k_n)}}\nabla{F\left({\bf x}_n^{(k_0,k_n-1)};\xi_i\right)}$.
        \ENDFOR
        \STATE Compute ${\mathbf x}_n^{(k_0,K_n)}-\hat{\mathbf x}^{(k_0)}$, and send $\mathbf Q\left({\mathbf x}_n^{(k_0,K_n)}-\hat{\mathbf x}^{(k_0)};s_n\right)$ to the server.\label{step:quantization}
    \ENDFOR
    \STATE The server computes $\Delta\hat{\mathbf x}^{(k_0)}$ according to:
    $\Delta\hat{\mathbf x}^{(k_0)}=\frac{1}{N}\sum_{n\in\mathcal N}\mathbf Q\left({\bf x}_n^{(k_0,K_n)}-\hat{\mathbf x}^{(k_0)};s_n\right)$,
    and sends $\mathbf Q(\Delta\hat{\mathbf x}^{(k_0)};s_0)$ to all $N$ workers.
\ENDFOR
\STATE The server and $N$ workers update $\hat{\mathbf x}^{(K_0+1)}$ according to: $
\hat{\mathbf x}^{(K_0+1)}:=\hat{\mathbf x}^{(K_0)}+\mathbf Q(\Delta\hat{\mathbf x}^{(K_0)};s_0)$,
and set $\mathbf x^*\left(\mathbf K,B\right)=\hat{\mathbf x}^{(K_0+1)}$.\label{step:global_average}
\end{algorithmic}
\end{algorithm}

\addtolength{\topmargin}{-0.02in}

\subsection{Convergence Analysis}
In the remaining of this paper, we assume that the following typical assumptions are satisfied \cite{YuHao}.
\begin{Asump}[I.I.D. Samples]\label{Asump:IID}
$\zeta_n,n\in\mathcal N$ are independent and identically distributed.
\end{Asump}

\begin{Asump}[Smoothness]\label{Asump:Smoothness}
For all $n\in\mathcal N$, $f_n({\bf x})$ is continuously differentiable, and its gradient is Lipschitz continuous, i.e., there exists a constant $L>0$ such that $\left\|\nabla{f_n({\mathbf x})}-\nabla{f_n({\mathbf y})}\right\|_2 \leq L\left\|{\mathbf x}-{\mathbf y}\right\|_2$, for all $\mathbf x,\mathbf y\in\mathbb R^D$.
\end{Asump}
\begin{Asump}[Bounded Variances]\label{Asump:BoundedVariances}
For all $n\in\mathcal N$, there exists a constant $\sigma>0$ such that $\mathbb E\left[\left\|\nabla{F\left({\mathbf x};\zeta_n\right)}-\nabla{f_n({\mathbf x})}\right\|_2^2\right]\leq{\sigma^2}$, for all $\mathbf x\in\mathbb R^D$.
\end{Asump}
\begin{Asump}[Bounded Second Moments]\label{Asump:BoundedSecondMoments}
For all $n\in\mathcal N$, there exists a constant $G>0$ such that $\mathbb E\left[\left\|\nabla{F\left({\mathbf x};\zeta_n\right)}\right\|_2^2\right] \leq {G^2}$, for all $ \mathbf x\in\mathbb R^D$.
\end{Asump}

For convenience of analysis, we define:
\vspace{-10pt}
\begin{align}
\tilde{\mathbf x}_n^{(k_0,k)}&\triangleq\left\{
    \begin{array}{ll}
    {\bf x}_n^{(k_0,k_n)},&k\in\mathcal K_n\\
    {\bf x}_n^{(k_0,K_n)},&k\in\mathcal K_{\max}\setminus\mathcal K_n
    \end{array},
\right.\nonumber\\
&{\quad\quad\quad}k_0\in\mathcal K_0, k\in\mathcal K_{\max}, n\in\mathcal N,\label{eq:local_update_redefine}\\
N_k&\triangleq\sum_{n\in\mathcal N}\mathbb I\left[k\leq K_n\right],\ k\in\mathcal K_{\max},\label{eq:N_k}\\
\bar{\mathbf x}^{(k_0,k)}&\triangleq\frac{1}{N}\sum_{n\in\mathcal N}\tilde{\mathbf x}_n^{(k_0,k)},\ k_0\in\mathcal K_0,k\in\mathcal K_{\max},\label{eq:x_avg}
\end{align}
where $\mathcal K_{\max}\triangleq\left\{1,2,\cdots,\max_{n\in\mathcal N}K_n\right\}$.
The goal is to synchronize the local iterations by letting each worker $n\in\mathcal N$ with $K_n<\max_{n\in\mathcal N}K_n$ run extra $\max_{n\in\mathcal N}K_n-K_n$ virtual local updates without changing its local model \cite{YuHao}.
Thus, $N_k$ in \eqref{eq:N_k} can be viewed as the number of workers conducting true local updates at the $k$-th synchronized local iteration within each global iteration,
and $\bar{\mathbf x}^{(k_0,k)}$ in \eqref{eq:x_avg} can be interpreted as the average of the local models at the $k$-th synchronized local iteration within the $k_0$-th global iteration.
The convergence of GenQSGD is summarized below.
\begin{Thm}[Convergence of GenQSGD]\label{Thm:Convergence_Q_n}
Suppose that Assumptions~\ref{Asump:Quantization},\ref{Asump:IID},\ref{Asump:Smoothness},\ref{Asump:BoundedVariances},\ref{Asump:BoundedSecondMoments} are satisfied and the step size $\gamma\in\left(0,\frac{1}{L}\right]$.
Then, for all $\mathbf K\in\mathbb Z_+^{N+1}$ and $B\in\mathbb Z_+$,
$\left\{\bar{\mathbf x}^{(k_0,k)}:\right.$ $\left.k_0\in\mathcal K_0,k\in\mathcal K_{\max}\right\}$
generated by GenQSGD satisfies:\footnote{We follow the convention in literature \cite{YuHao} to use the expected squared gradient norm to characterize the convergence given limited realizations.}
\begin{align}\label{eq:UpBound_Q_n}
&\frac{1}{K_0\frac{1}{N}\sum_{n\in\mathcal N}{K_n}}\sum_{k_0\in\mathcal K_0}{\sum_{k\in\mathcal K_{\max}}{\frac{N_k}{N}\mathbb E\left[\left\|\nabla f\left(\bar{\mathbf x}^{(k_0,k-1)}\right)\right\|_2^2\right]}}\nonumber\\
\leq&\frac{2}{\gamma K_0\frac{1}{N}\sum_{n\in\mathcal N}{K_n}}\left(f\left(\hat{\mathbf x}^{(1)}\right)-f^*\right)+4\gamma^2\max_{n\in\mathcal N}K_n^2G^2L^2\nonumber\\
&+\frac{L\gamma\sigma^2}{NB}+\frac{2L\gamma G^2\sum_{n\in\mathcal N}(q_{s_0}+q_{s_n}+q_{s_0}q_{s_n})K_n^2}{\sum_{n\in\mathcal N}K_n},
\end{align}
where $f^*$ is the optimal value of Problem~\ref{Prob1}.
\end{Thm}
\begin{Proof}[Sketch]
Under Assumption~\ref{Asump:Quantization} (i) and Assumption~\ref{Asump:Smoothness}, by extending \cite[Lemma 6]{FedPAQ}, we can show $\mathbb E\left[f\left(\hat{\mathbf x}^{(k_0+1)}\right)\right]\leq\mathbb E\left[f\left(\bar{\mathbf x}^{(k_0, K_{\max})}\right)\right]+\frac{L}{2}\mathbb E\left[\left\|\hat{\mathbf x}^{(k_0+1)}-\bar{\mathbf x}^{(k_0, K_{\max})}\right\|_2^2\right]$,
where $K_{\max}\triangleq\max_{n\in\mathcal N}K_n$.
Under Assumptions~\ref{Asump:IID}, \ref{Asump:Smoothness}, \ref{Asump:BoundedVariances}, and \ref{Asump:BoundedSecondMoments}, by extending the proof of \cite[Theorem 3]{YuHao}, we can show $\mathbb E\left[f\left(\bar{\mathbf x}^{(k_0, K_{\max})}\right)\right]\leq\mathbb E\left[f\left(\hat{\mathbf x}^{(k_0)}\right)\right]-\frac{\gamma}{2}\sum_{k\in\mathcal K_{\max}}\frac{N_k}{N}\mathbb E\left[\left\|\nabla f\left(\bar{\mathbf x}^{(k_0, k-1)}\right)\right\|_2^2\right]+\left(2\gamma^3K_{\max}^2G^2L^2+\frac{L\gamma^2\sigma^2}{2NB}\right)\sum_{k\in\mathcal K_{\max}}\frac{N_k}{N}$.
Under Assumptions~\ref{Asump:Quantization} and \ref{Asump:BoundedSecondMoments}, by extending the proof of \cite[Lemma 8]{FedPAQ}, we can show $\mathbb E\left[\left\|\hat{\mathbf x}^{(k_0+1)}-\bar{\mathbf x}^{(k_0, K_{\max})}\right\|_2^2\right]\leq\frac{2\gamma^2G^2}{N}\sum_{n\in\mathcal N}\left(q_{s_0}+q_{s_n}+q_{s_0}q_{s_n}\right)K_n^2$.
From the above three inequalities, we can prove Theorem~\ref{Thm:Convergence_Q_n}.
\end{Proof}
\begin{Rmk}[Interpretation of Theorem~\ref{Thm:Convergence_Q_n}]
Theorem~\ref{Thm:Convergence_Q_n} indicates that the convergence of GenQSGD is influenced by the algorithm parameters $\mathbf K,B$ and the quantization parameters $\mathbf s$.
In the following, we illustrate how the four terms on the R.H.S of \eqref{eq:UpBound_Q_n} change with $\mathbf K$, $B$ and $\mathbf s$.
The first term decreases with $K_0$ and vanishes as $K_0\rightarrow\infty$.
For any $n\in\mathcal N$, the first term decreases with $K_n$, and the last term increases with $K_n$.
Fixing $\sum_{n\in\mathcal N}K_n$, the second term and the last term increase with $\max_{n\in\mathcal N}K_n$.
The third term decreases with $B$ due to the decrease of the variance of the stochastic gradients.
For all $n\in\setNbar$, the last term increases with $q_{s_n}$ (decreases with $s_n$) and vanishes as $s_n\rightarrow\infty$.
\end{Rmk}
\begin{Rmk}[Generality of Convergence of GenQSGD]
Theorem~\ref{Thm:Convergence_Q_n} for the convergence of GenQSGD with $q_{s_n}=0\ (s_n=\infty),n\in\setNbar$ and $B=1$ reduces to the convergence result of PR-SGD in \cite[Theorem 3]{YuHao}.
Note that \cite{YuHao} does not consider quantization at the server and
workers and utilises only one local sample in each local iteration.
\end{Rmk}

\section{Performance Metrics}\label{Sec:PerformanceMetrics}
In this section, we introduce the performance metrics for implementing GenQSGD in the edge computing system, including the time cost, energy cost, and convergence error.

\subsection{Time Cost}
For each worker $n\in\mathcal N$, let $C_n$ denote the number of CPU-cycles required to compute $\nabla F({\mathbf x};\xi_i)$ for all $\mathbf x\in\mathbb R^{D}$ and $\xi_i\in\mathcal I_n$.
Then, the computation time (seconds) for worker $n$ to execute $K_n$ local iterations is given by $\frac{BK_nC_n}{F_n}$.
As the $N$ workers execute local iterations in a parallel manner,
within each global iteration, the computation time for local model training at the workers is $B\max_{n\in\mathcal N}\frac{K_nC_n}{F_n}$.
Let $C_0$ denote the number of CPU-cycles required for the server to compute the average of the received quantized local model updates from all $N$ workers.
Then, the computation time for global averaging at the server within each global iteration is given by $\frac{C_0}{F_0}$.
Thus, the communication time for worker $n$ to transmit $\mathbf Q({\mathbf x}_n^{(k_0,K_n)}-\hat{\mathbf x}^{(k_0)};s_n)$ to the server is given by $\frac{M_{s_n}}{r_n}$, for all $k_0\in\mathcal K_0$.
As the $N$ workers transmit their quantized messages synchronously at given transmission rates using FDMA, within each global iteration, the communication time for sending quantized local updates from all $N$ workers to the server is $\max_{n\in\mathcal N}\frac{M_{s_n}}{r_n}$.
Thus, the communication time for the server to multicast $\mathbf Q(\Delta\hat{\mathbf x}^{(k_0)};s_0)$ to all $N$ workers within the $k_0$-th global iteration is given by $\frac{M_{s_0}}{r_0}$, for all $k_0\in\mathcal K_0$.
The time cost for implementing GenQSGD in the edge computing system, denoted by $T(\mathbf K,B)$, is defined as the sum of the overall computation time and communication time.
As GenQSGD has $K_0$ global iterations, we have:
\begin{align}\label{eq:Time}
&T(\mathbf K,B)=\nonumber\\
&K_0\left(B\max_{n\in\mathcal N}\frac{C_n}{F_n}K_n+\frac{C_0}{F_0}+\max_{n\in\mathcal N}\frac{M_{s_n}}{r_n}+\frac{M_{s_0}}{r_0}\right).
\end{align}

\subsection{Energy Cost}
The computation energy for worker $n\in\mathcal N$ to execute $K_n$ local iterations is given by $\alpha_n BK_nC_n{F_n}^2$, where $\alpha_n$ is a constant factor determined by the switched capacitance of worker $n$.
Then, within each global iteration, the computation energy for local model training at the workers is $B\sum_{n\in\mathcal N}\alpha_nK_nC_n{F_n}^2$.
Similarly, the computation energy for global averaging at the server within each global iteration is given by $\alpha_0 C_0{F_0}^2$,
where $\alpha_0$ is a constant factor determined by the switched capacitance of the server.
The communication energy for worker $n$ to transmit $\mathbf Q({\mathbf x}_n^{(k_0,K_n)}-\hat{\mathbf x}^{(k_0)};s_n)$ to the server is given by $\frac{p_n M_{s_n}}{r_n}$, for all $k_0\in\mathcal K_0$.
Analogously, the communication energy for sending quantized local updates from all $N$ workers to the server is $D\sum_{n\in\mathcal N}\frac{p_n\log_2s_n}{r_n}$.
Analogously, the communication energy for the server to multicast $\mathbf Q(\Delta\hat{\mathbf x}^{(k_0)};s_0)$ to all $N$ workers at the $k_0$-th global iteration is given by $\frac{p_0M_{s_0}}{r_0}$, for all $k_0\in\mathcal K_0$.
The energy cost for implementing GenQSGD in the edge computing system, denoted by $E(\mathbf K,B)$, is defined as the sum of the overall computation energy and communication energy.
As GenQSGD has $K_0$ global iterations, we have:
\begin{align}\label{eq:Energy}
&E(\mathbf K,B)=\nonumber\\
&K_0\left(B\sum_{n\in\mathcal N}\alpha_nC_nF_n^2K_n+\alpha_0C_0F_0^2+\sum_{n\in\setNbar}\frac{p_nM_{s_n}}{r_n}\right).
\end{align}

\subsection{Convergence Error}
The upper bound on the convergence error of GenQSGD in Theorem~\ref{Thm:Convergence_Q_n} relies on $f^*$, which is usually unknown before solving Problem~\ref{Prob1}.
However, in some cases, we can estimate $f^*$ or bound $f^*$ from below.
For tractability, in the remaining of the paper, we define the convergence error (upper bound), denoted by $C(\mathbf K,B)$, as:
\begin{align}\label{eq:convergence}
C(\mathbf K,B)&=\frac{c_1}{\sum_{n\in\mathcal N}K_0K_n}+c_2\max_{n\in\mathcal N}K_n^2+\frac{c_3}{B}\nonumber\\
&+\frac{2L\gamma G^2\sum_{n\in\mathcal N}(q_{s_0}+q_{s_n}+q_{s_0}q_{s_n})K_n^2}{\sum_{n\in\mathcal N}K_n},
\end{align}
where $c_1\triangleq\frac{2N\left(f\left(\hat{\mathbf x}^{(1)}\right)-\delta\right)}{\gamma}$ with $\delta$ being a known estimate of $f^*$ or a lower bound of $f^*$,
$c_2\triangleq4\gamma^2G^2L^2$,
and $c_3\triangleq\frac{L\gamma\sigma^2}{N}$.

\section{Optimization of Algorithm Parameters}\label{sec:Opt_fix}
In this section, we formulate and solve the optimization problem of the algorithm parameters of GenQSGD according to the parameters of the underlying edge computing system and machine learning problem.
\subsection{Problem Formulation}\label{SubSec:Fixed_Q}

For tractability, in the optimization, we relax the integer constraints, $\mathbf K\in\mathbb Z_+^{N+1}$ and $B\in\mathbb Z_+$, to their continuous counterparts, $\mathbf K\succ\mathbf 0$ and $B>0$, respectively.
We can easily construct an integer point based on a continuous point with similar performance.
Specifically, we optimize the algorithm parameters of GenQSGD to minimize energy cost $E(\mathbf K,B)$ subject to the time constraint and convergence error constraint.

\begin{Prob}[Optimization Problem]\label{Prob:fix}
\begin{align}
E^*\triangleq\min_{\mathbf K\succ\mathbf 0, B>0}&{\quad}E(\mathbf K,B)\nonumber\\
\mathrm{s.t.}&{\quad}T(\mathbf K,B)<T_{\max},\label{eq:Cons_time}\\
&{\quad}C(\mathbf K,B)<C_{\max},\label{eq:Cons_conv}
\end{align}
where $E(\mathbf K,B)$, $T(\mathbf K,B)$ and $C(\mathbf K,B)$ are given by \eqref{eq:Energy}, \eqref{eq:Time} and \eqref{eq:convergence} respectively, and $T_{\max}$ and $C_{\max}$ denote the limits on the time cost and convergence error, respectively.
\end{Prob}
Apparently, the minimum energy cost $E^*$ does not increase (usually decreases) with the limits on time cost and convergence error, i.e., $T_{\max}$ and $C_{\max}$, indicating an optimal tradeoff among the energy cost, time cost, and convergence error.
Similar problems can be formulated in terms of the three performance metrics to optimize the tradeoff.
Later, we shall see that the proposed optimization method for Problem~\ref{Prob:fix} can be readily extended to handle those formulations.
From \eqref{eq:Time}, \eqref{eq:Energy}, and \eqref{eq:convergence}, we know that the parameters of Problem~\ref{Prob:fix} include the system parameters, i.e.,
$F_n,C_n,\alpha_n,p_n,r_n,s_n,n\in\setNbar$, which are known to the server, and the parameters of the expected risk function $f$, i.e., $L$, $\sigma$, $G$, and $\delta$, which can be obtained by the server from pre-training based on the data stored on the server.
Thus, Problem~\ref{Prob:fix}, assumed to be feasible, can be solved by the server (in an offline manner) before implementing GenQSGD in the edge computing system.

The constraints in \eqref{eq:Cons_time} and \eqref{eq:Cons_conv} are non-convex and contain non-differentiable functions, and the objective function is non-convex.
Thus, Problem~\ref{Prob:fix} is a challenging non-convex problem with non-differentiable constraint functions.\footnote{There are in general no effective methods for solving a non-convex problem optimally.
The goal of solving a non-convex problem is usually to obtain a KKT point.}

\subsection{Solution}\label{SubSub:fix_solution}
In this part, we develop an algorithm to obtain a KKT point of an equivalent problem of Problem~\ref{Prob:fix}.
First, we address the challenge caused by the non-differentiable constraint functions in \eqref{eq:Cons_time} and \eqref{eq:Cons_conv}.
Specifically, we introduce auxiliary variables $T_1>0$ and $T_2>0$, replace $\max_{n\in\mathcal N}\frac{C_n}{F_n}K_n$ in \eqref{eq:Cons_time} and $\max_{n\in\mathcal N}K_n^2$ in \eqref{eq:Cons_conv} with $T_1$ and $T_2^2$, respectively, and add the following inequality constraints:
\begin{align}
\frac{C_n}{F_n}K_n T_1^{-1}\leq1,{\quad}n\in\mathcal N,\label{eq:fix_epi_cons1}\\
K_nT_2^{-1}\leq1,{\quad}n\in\mathcal N.\label{eq:fix_epi_cons2}
\end{align}
Then, we can transform Problem~\ref{Prob:fix} into the following problem with differentiable constraint functions.
\begin{Prob}[Equivalent Problem of Problem~\ref{Prob:fix}]\label{Prob:fix_epi}
\begin{align}
&\min_{\substack{\mathbf K\succ\mathbf0,\\B,T_1,T_2>0}}{\quad}E(\mathbf K,B)\nonumber\\
&\mathrm{s.t.}{\quad}\eqref{eq:fix_epi_cons1},\ \eqref{eq:fix_epi_cons2},\nonumber\\
&\left(\frac{C_0}{F_0}+\max_{n\in\mathcal N}\frac{M_{s_n}}{r_n}+\frac{M_{s_0}}{r_0}\right)K_0+BK_0T_1\leq T_{\max},\label{eq:fix_epi_cons_T}
\end{align}
\begin{align}
&\frac{c_1}{\sum_{n\in\mathcal N}K_0K_n}+c_2T_2^2+\frac{c_3}{B}\nonumber\\
&+\frac{2L\gamma G^2\sum_{n\in\mathcal N}(q_{s_0}+q_{s_n}+q_{s_0}q_{s_n})K_n^2}{\sum_{n\in\mathcal N}K_n}\leq C_{\max}.\label{eq:fix_epi_cons_C}
\end{align}
\end{Prob}
By contradiction, we can easily show that Problem~\ref{Prob:fix} and Problem~\ref{Prob:fix_epi} are equivalent.
Note that $E(\mathbf K,B)$ and the constraint function in \eqref{eq:fix_epi_cons_T} are posynomials, the constraint functions in \eqref{eq:fix_epi_cons1} and \eqref{eq:fix_epi_cons2} are monomials, the first term of the constraint function in \eqref{eq:fix_epi_cons_C} is a ratio between a constant and a posynomial, and the last term of the constraint function in \eqref{eq:fix_epi_cons_C} is a ratio between two posynomials.
Thus, Problem~\ref{Prob:fix_epi} is a non-convex problem that is more complicated than a CGP \cite{CGP}.

In the following, using GIA \cite{GIA} and the tricks for solving CGP \cite{CGP}, we propose an iterative algorithm, i.e. Algorithm~\ref{Alg:fix}, to obtain a KKT point of Problem~\ref{Prob:fix_epi}.
The idea is to solve a sequence of successively refined approximate geometric programs (GPs), each of which is obtained by lower bounding the denominators of the first term and the last term of the constraint function in \eqref{eq:fix_epi_cons_C} using the arithmetic-geometric mean inequality.
Specifically, at iteration $t$, update $\left(\mathbf K^{(t)}, B^{(t)}\right)$ by solving the following approximate GP of Problem~\ref{Prob:fix_epi}, which is parameterized by $\mathbf K^{(t-1)}$ obtained at iteration $t-1$.
\begin{algorithm}[t]
\caption{Algorithm for Obtaining a KKT Point of Problem~\ref{Prob:fix_epi}}
\label{Alg:fix}
\begin{algorithmic}[1]
\STATE {\bf{Initialize:}} Choose any feasible solution $\left(\mathbf K^{(0)}, B^{(0)}, T_1^{(0)}, T_2^{(0)}\right)$ of Problem~\ref{Prob:fix_GP} as the initial point, and set $t=1$.
\REPEAT
    \STATE Compute $\left(\mathbf K^{(t)}, B^{(t)}, T_1^{(t)}, T_2^{(t)}\right)$ by transforming Problem~\ref{Prob:fix_GP} into a GP in convex form, and solving it with standard convex optimization techniques.
    \STATE Set $t:=t+1$.
\UNTIL{Some convergence criteria is met.}
\end{algorithmic}
\end{algorithm}

\begin{figure*}[t]
\begin{center}
\subfigure[\small{The training loss and test accuracy of GenQSGD versus the convergence error limit $C_{\max}$ at $T_{\max}=1500$.}]
{\resizebox{5.57cm}{!}{\includegraphics{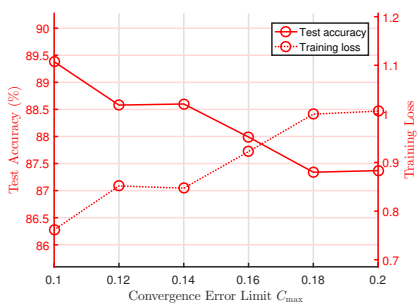}}\label{Fig:loss_acc_vs_convergence}}\quad\quad
\subfigure[\small{The energy cost versus the convergence error limit $C_{\max}$ at $T_{\max}=1500$.}]
{\resizebox{5.57cm}{!}{\includegraphics{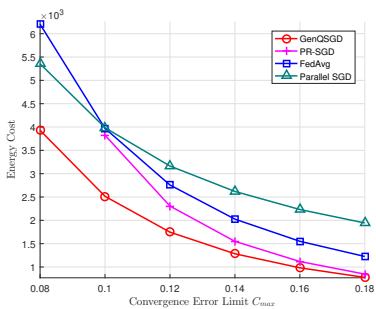}}\label{Fig:Energy_vs_Cmax}}\quad\quad
\subfigure[\small{The energy cost versus the time limit $T_{\max}$ at $C_{\max}=0.1$.}]
{\resizebox{5.57cm}{!}{\includegraphics{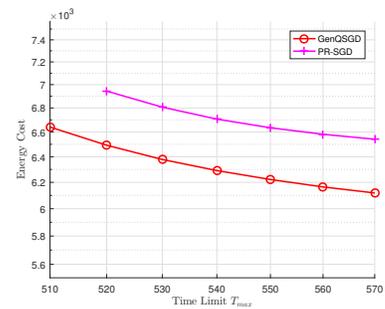}}\label{Fig:Energy_vs_Tmax}}
\end{center}
\vspace{-2mm}
\caption{\small{The performance of GenQSGD. Some points for the curves of the baseline algorithms are missing, as their parameter optimization problems are not feasible at some $C_{\max}$ and $T_{\max}$.}}
\vspace{-2mm}
\label{fig:performance_comparison}
\end{figure*}

\begin{Prob}[Approximate GP at Iteration $t$]\label{Prob:fix_GP}
\begin{align}
&\min_{\substack{\mathbf K\succ\mathbf0,\\B,T_1,T_2>0}}{\quad}E(\mathbf K,B)\nonumber\\
&\mathrm{s.t.}{\quad}\eqref{eq:fix_epi_cons1},\ \eqref{eq:fix_epi_cons2},\
\eqref{eq:fix_epi_cons_T},\nonumber\\
&\frac{c_1}{K_0\prod_{n\in\mathcal N}\left(\frac{K_n}{\beta_n^{(t-1)}}\right)^{\beta_n^{(t-1)}}}
+c_2T_2^2+\frac{c_3}{B}\nonumber\\
+&\frac{2L\gamma G^2\sum_{n\in\mathcal N}(q_{s_0}+q_{s_n}+q_{s_0}q_{s_n})K_n^2}{\prod_{n\in\mathcal N}\left(\frac{K_n}{\beta_n^{(t-1)}}\right)^{\beta_n^{(t-1)}}}\leq C_{\max},\label{eq:fix_epi_cons_C_approx}
\end{align}
where $\beta_n^{(t-1)}\triangleq\frac{K_n^{(t-1)}}{\sum_{n\in\mathcal N}K_n^{(t-1)}}$.
Let $\left(\mathbf K^{(t)}, B^{(t)}, T_1^{(t)}, T_2^{(t)}\right)$ denote an optimal solution of Problem~\ref{Prob:fix_GP}.
\end{Prob}
The constraint function in \eqref{eq:fix_epi_cons_C_approx}, which is an approximation of the constraint function in \eqref{eq:fix_epi_cons_C} at $\mathbf K^{(t-1)}$, is a posynomial.
As a result, Problem~\ref{Prob:fix_GP} is a standard GP and can be readily transformed into a convex problem and solved by using standard convex optimization techniques.
The details are summarized in Algorithm~\ref{Alg:fix}.
Following \cite[Proposition 3]{CGP}, we have the following result.
\begin{Thm}[Convergence of Algorithm~\ref{Alg:fix}]\label{Thm:fix_convergence}
$\left(\mathbf K^{(t)},B^{(t)},T_1^{(t)},T_2^{(t)}\right)$ obtained by Algorithm~\ref{Alg:fix} converges to a KKT point of Problem~\ref{Prob:fix_epi}, as $t\rightarrow\infty$.
\end{Thm}
\begin{Proof}[Sketch]
By \cite[Proposition 3]{CGP}, we know that Properties (i), (ii), and (iii) in \cite{GIA} are satisfied.
As $\left(\mathbf K^{(0)}, B^{(0)}, T_1^{(0)}, T_2^{(0)}\right)$ is feasible, Problem~\ref{Prob:fix_GP} is feasible and hence has zero duality gap \cite{GP}.
Therefore, by \cite[Theorem 1]{GIA}, we can show Theorem~\ref{Thm:fix_convergence}.
\end{Proof}

\section{Numerical Results}

In this section, we evaluate the performance of GenQSGD in Algorithm~\ref{Alg:GenQSGD} with the algorithm parameters $\mathbf K$ and $B$ obtained by Algorithm~\ref{Alg:fix}. We consider a ten-class classification problem with MNIST dataset ($I=6\times10^4$). We partition the MNIST training dataset into $N=10$ subsets, each of which is stored on one worker. We consider a three-layer neural network, including an input layer composed of $784$ cells, a hidden layer composed of $128$ cells, and an output layer composed of $10$ cells. Thus, $D=101632$. We use the sigmoid activation function for the hidden layer and the softmax activation function for the output layer. We consider the cross entropy loss function. We choose
$\alpha_n=2\times10^{-28},n\in\setNbar$,
$F_0=3\times10^9$ (cycles/s),
$C_0=1000$ (cycles),
$p_0=20$ (W),
$r_0=7.5\times10^7$ (b/s),
$F_n\in[2.7\times10^8,2.7\times10^9]$ (cycles/s), $n\in\mathcal N$,
$C_n\in[1.8\times10^7,1.8\times10^8]$ (cycles), $n\in\mathcal N$,
$p_n\in[0.27,2.7]$ (W), $n\in\mathcal N$,
$r_n\in[9\times10^5,9\times10^6]$ (b/s), $n\in\mathcal N$,
$q_{s_0}=4.9$,
$q_{s_n}=9.9,n\in\mathcal N$,
and $\gamma=0.03$.
We set $L=0.034$, $\sigma=18$, and $G=0.6$, which are obtained by pre-training. We also apply three baseline algorithms, namely, PR-SGD \cite{YuHao}, FedAvg \cite{FedAvg}, and parallel SGD (P-SGD) \cite{YuHao}, with the algorithm parameters obtained using the proposed optimization framework, to the considered edge computing system.

Fig.~\ref{Fig:loss_acc_vs_convergence} shows the training loss and test accuracy of GenQSGD versus the convergence error limit $C_{\max}$.
Fig.~\ref{Fig:loss_acc_vs_convergence} indicates that by imposing the convergence error constraint in \eqref{eq:Cons_conv}, we can effectively control the training loss and test accuracy.
Fig.~\ref{Fig:Energy_vs_Cmax} and Fig.~\ref{Fig:Energy_vs_Tmax} show the energy cost versus the convergence error limit $C_{\max}$ and the time limit $T_{\max}$, respectively.
Note that some points for the curves of the baseline algorithms are missing, as their parameter optimization problems are not feasible at some $C_{\max}$ and $T_{\max}$.
From Fig.~\ref{Fig:Energy_vs_Cmax} and Fig.~\ref{Fig:Energy_vs_Tmax}, we can see that the energy cost of each feasible algorithm decreases with $C_{\max}$ and $T_{\max}$, reflecting a tradeoff among the energy cost, time cost, and convergence error.
GenQSGD significantly outperforms the baseline algorithms, indicating the importance of designing a general FL algorithm and optimally adapting its algorithm parameters to the underlying edge computing system and machine learning problem.

\section{Conclusion}
In this paper, we presented a general quantized parallel mini-batch SGD algorithm for FL and analyzed its convergence error when applied to a general unconstrained machine learning problem. Then, we introduced a rigorous framework for optimizing the algorithm parameters to achieve an optimal tradeoff among the energy cost, time cost, and convergence error. Numerical results reveal the significance of optimally designing FL algorithms in practical edge computing systems.

\normalem
\bibliographystyle{IEEEtran}      
\bibliography{GenQSGD_GC_final}                        

\clearpage

\end{document}